\documentclass[lettersize,journal]{IEEEtran}
\usepackage{amsmath,amsfonts}
\usepackage{array}
\usepackage[caption=false,font=normalsize,labelfont=sf,textfont=sf]{subfig}
\usepackage{textcomp}
\usepackage{stfloats}
\usepackage{url}
\usepackage{verbatim}
\usepackage{graphicx}
\usepackage{cite}
\usepackage{algorithm2e}
\usepackage{amsmath,amsthm,amssymb} 
\newtheoremstyle{cited}%
{3pt}
{3pt}
{}
{1pc}
{\bfseries}
{\textbf{:}}
{.5em}
{\thmname{#1} \thmnumber{#2}\thmnote{\normalfont#3}}

\theoremstyle{cited}

\usepackage{amsmath}

\newtheorem{definition}{Definition}

\begin{document}

\title{Geometry-Aware Coverage Path Planning for Depowdering on Complex 3D Surfaces}

\author{Van-Thach Do and Quang-Cuong Pham
\thanks{This study is supported under the RIE2020 Industry Alignment Fund –Industry Collaboration Projects (IAF-ICP) Funding Initiative, as well as cash and in-kind contribution from the industry partner, HP Inc., through the HP-NTU Digital Manufacturing Corporate Lab. (Corresponding author: Van-Thach Do)
	
	The authors are with the HP-NTU Digital Manufacturing Corporate Lab and School of Mechanical and Aerospace Engineering, Nanyang Technological University, Singapore. {\tt\small \{thach.do,cuong\}@ntu.edu.sg}. }}



\maketitle

\begin{abstract}
This paper presents a new approach to obtaining nearly complete coverage paths (CP) with low overlapping on 3D general surfaces using mesh models. The CP is obtained by segmenting the mesh model into a given number of clusters using constrained centroidal Voronoi tessellation (CCVT) and finding the shortest path from cluster centroids using the geodesic metric efficiently. We introduce a new cost function to harmoniously achieve uniform areas of the obtained clusters and a restriction on the variation of triangle normals during the construction of CCVTs. The
obtained clusters can be used to construct high-quality viewpoints (VP) for visual coverage tasks. Here, we utilize the planned VPs as cleaning configurations to perform residual powder removal in additive manufacturing using manipulator robots. The self-occlusion of VPs and ensuring collision-free robot configurations are addressed by integrating a proposed optimization-based strategy to find a set of candidate rays for each VP into the motion planning phase. CP planning benchmarks and physical experiments are conducted to demonstrate the effectiveness of the proposed approach. We show that our approach can compute the CPs and VPs of various mesh models with a massive number of triangles within a reasonable time.
\end{abstract}


\section{Introduction}
	\label{sec:introduction}
\IEEEPARstart{C}{overage} path planning (CPP) is the problem of computing a path that traverses all points in a given domain. Several criteria for a CPP include complete coverage, no overlap, and satisfying task-based additional requirements \cite{galceran2013survey}. Designing a CPP algorithm to satisfy all those criteria is challenging and may not be achieved in practice. Thus, there will be a tradeoff in selecting the priorities among those criteria to meet the quality of a specific task.

Numerous previous CPP approaches aim to generate polylines with a nearly constant swath width on the target surface to ensure complete and consistent coverage. The coverage paths can be utilized for various tasks, such as indoor cleaning, surveillance, agriculture services, etc. Additionally, CPP plays a pivotal role in perception-based surface treatments using robotic end-effectors, including inspections and 3D reconstructions, where the quality of the selected viewpoints (VP) significantly affects the performance. Most existing methods for selecting VPs either randomly generate a set of candidate VPs around the object or choose them based on the normal direction of each triangle face of the mesh \cite{PEUZINJUBERT2021103094}. While these methods may perform well for simple meshes, they can lead to inadequate coverage of complex geometries and high overlap and may increase computation time if the number of sample VPs is high. Furthermore, because these methods do not account for self-collision or collisions between the robot and the surrounding objects, they cannot ensure that visual sensors reach the object surface.

In contrast to prior approaches, this study presents a new CPP algorithm that can attain nearly complete coverage and low overlapping on the object surface while producing high-quality VPs for cleaning, model-based inspection, or 3D reconstruction tasks. We also address the self-occlusion (occlusion between VPs and other triangle faces in the mesh) and the collision between the robot body and the surrounding environment. Then, we leverage the obtained VPs for an industrial cleaning application using manipulator robots. Our approach consists of three main contributions as follows:

\begin{figure}[t]\centering
	\includegraphics[width=0.49\textwidth]{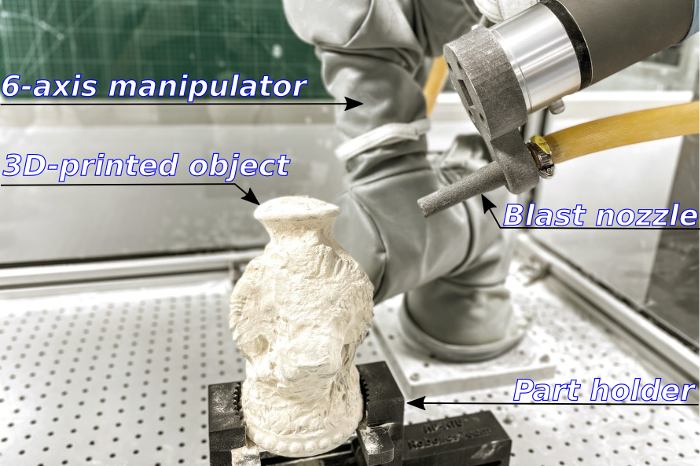}
	\caption{\label{fig_sim_env}Experimental setup of a depowdering system for 3D-printed objects.}
\end{figure}

(i) We present a new energy function for the constrained centroidal Voronoi tessellation (CCVT) method \cite{nguyen2009constrained} to segment the mesh model $\mathbb{M}$ into $m$ clusters with low standard deviations (SD) in both area and triangle normal. VPs are constructed based on the mass centroids and proxy normals \cite{skrodzki2020variational} (average triangle normals) of the resulting clusters.

(ii) We propose an efficient approach, namely geodesic decomposition calculation, that has low time complexity to calculate the exact geodesic distances (GD) between centroids. Then, the coverage path on the surface is computed based on the obtained distance costs between centroids.

(iii) Compared to prior works, we address both issues that may be encountered when positioning the robot tool at the obtained VPs: (1) self-occlusion between rays at VPs and other triangles in $\mathbb{M}$; and (2) infeasible robot configurations at computed poses due to joint limits or collisions with surrounding objects (rays at VPs in these cases are referred to as invalid rays). An efficient correction algorithm is proposed to obtain valid robot configurations with the closest direction (if possible) to the invalid rays by utilizing a set of candidate rays formed by an optimization strategy. We apply the proposed method to automate the removal of residual powder from 3D-printed (3DP) parts with complex geometries, which, to the best of our knowledge, has not been done before.

\subsection{Related Work}
The CPP problem has been widely studied and applied to various robotic applications \cite{gleeson2022generating, bormann2018indoor,almadhoun2019survey,hameed2014intelligent,tan2021Review}, where most of the target surfaces are flat or have low curvature geometries. 
In \cite{lin2017robot}, a global surface parameterization-based CPP is proposed to ensure non-intersecting paths with great coverage for general surfaces with complex topology. By considering minimizing energy consumption caused by gravity, a CPP algorithm is proposed to plan the Fermat spiral paths for mobile robots on general terrain surfaces \cite{wu2019energy}. In those approaches, the CPP results in nearly complete coverage paths on the target surface.

CPP-based surface cleaning solutions using robot manipulators have been widely developed. In  \cite{moura2018automation}\cite{kabir2017automated}, surfaces cleaning applications are proposed, where the coverage paths are generated by projecting from 2D to 3D and obtaining from intersecting with equally spaced parallel planes, respectively. However, the quality of the generated coverage paths may degrade on high-curvature surfaces due to distortion. In \cite{Hess2012}, the problem of minimizing the cost function in the joint space for surface cleaning tasks is addressed. Here, the object surface is modeled as a set of planar patches. However, the obtained robot configurations may yield high overlapping rates due to randomly picking points from the acquired point cloud during the construction of surface patches. Also, the issues of self-occlusion and infeasible configurations due to joint limits are not addressed. 
In \cite{mcgovern2022uv}, a new method is introduced to generate a uv grid on three types of 3D freeform surfaces formed by non self-intersecting freeform curves. The method aims to create even coverage paths while considering task constraints on the end-effector pose. However, this approach necessitates human operators to provide transformed axes, and the resulting task constraints may not be suitable for high curvature surfaces, where the self-occlusion phenomenon exists.

In 3D inspection and reconstruction tasks, VP planning plays a crucial role in determining the optimal sets of sensor poses. The optimal set refers to the minimum number of VPs required to capture the target objects or scenes at a high coverage rate. In model-based methods, the VPs can be obtained in different ways. VPs are randomly generated \cite{Dornhege2013,jing2017model}. For high-curvature meshes, a large number of VP samples is required to ensure complete coverage. In \cite{Gronle_2016,englot2012sampling}, VPs are sampled from mesh vertices. This approach suffers from VP redundancy.
Another VP generation method is patch sampling \cite{prieto2003cad}, which involves segmenting the mesh model into sub-regions under specified constraints. The most relevant to our work is \cite{mosbach2021feature}, where the mesh model is segmented using B-splines and guided by a set of feature functionals. Due to its independence from mesh resolution, this approach can lead to a reduction in the number of VPs compared to random sampling methods. However, the computed VPs in the \textit{subdivided segments} step heavily rely on processing B-spline surfaces. This may result in a large number of redundant VPs at intersections of high-curvature B-spline surfaces. Also, the self-occlusion of VPs is not handled in that study. In \cite{glorieux2020coverage}, a targeted VP sampling strategy is proposed to find the optimal next VP. This is done by iteratively reformulating and solving the search problem, aiming to minimize the number of VPs and the travel cost while ensuring robot kinematics and avoiding collisions. However, the iterative search and sampling of every triangle face can result in a long computation time for high-resolution meshes.
Our proposed method, which segments the mesh globally and derives VPs from the obtained clusters, benefits from the efficient computation of patch sampling while achieving nearly uniform areas of the obtained clusters for low curvature meshes. Additionally, we address the self-occlusion of VPs and compute an optimal set of collision-free robot configurations for performing the task. 

Additive manufacturing is popular due to its ability to fabricate complex objects rapidly from CAD models. 
However, post-processing 3DP parts, such as support and powder removal and surface finishing, is time-consuming and expensive and poses potential hazards to workers. To assist and automate these processes, vision systems have been proposed to identify 6D poses of the 3DP parts under partly occluded by powder \cite{lim2021automated,nguyen2020development,Liu2022Depowdering}. In \cite{nguyen2020development}, a full pipeline is proposed to perform residual powder removal. However, the designed mechanism aims to clean flat or low-curvature objects. In \cite{Liu2022Depowdering}, a vision-based approach is proposed to perform depowdering for 3DP parts located in a powder bed. However, replacing actual powder materials, which are highly adhesive, with children's play sand simplifies this work compared to real-world scenarios. In their work, the roll and pitch motions of the robot tool are maintained during the operation, limiting the ability to handle complex objects. Their future work, i.e., investigating an advanced path planning algorithm to improve the depowdering automation, is addressed in our study.
\subsection{Automated Depowdering Platform for 3DP Parts}
Fig. \ref{fig_sim_env} shows the experimental cleaning system for 3DP parts, which consists of a 6-axis robotic manipulator, a blast nozzle tool, a part holder, and a 3DP part. The blast nozzle is attached to the robot end-effector, and a high-pressure compressed air stream containing tiny glass beads flows through the nozzle head to remove residual powder. Our goal is to propose an efficient CPP algorithm with collision-free robot trajectories to automate the removal of residual powder from 3DP parts after unpacking from 3D printing stations.
\section{Method}
\subsection{Computation of low curvature clusters using CCVT method}\label{sec_3a}
	Let $\mathbb{M} = \{\mathbb{F},\mathbb{N}\}$ be the input mesh, where $\mathbb{F}$ and $\mathbb{N}$ are the set of triangle faces and their corresponding normal vectors, respectively. Our objective is to divide $\mathbb{M}$ into $m$ connected, low curvature clusters of faces $\{\mathbb V_i\}^m_{i=1}$ by computing discrete CCVT on $\mathbb{M}$ with triangle normal constraints.
\begin{definition}
	Discrete Voronoi tessellation. 
	Given a discrete set of points $\mathbb{W} = \{\mathbf y_i\}^{n_t}_{i=1} \subset \Re^N$, where $\mathbf y_i$ are centroids of $n_t$ faces in $\mathbb F$. A set $\{\mathbb V_i\}^m_{i=1}$ is a tessellation of $\mathbb W$ if $\mathbb V_i \cap \mathbb V_j = \emptyset, \forall i\neq j$ and $\cup^k_{i=1}\mathbb V_i=\mathbb W$. Voronoi sets are defined as
	\begin{eqnarray}
		\mathbb V_i = \{ \mathbf y \in \mathbb W|\lVert\mathbf y-\mathbf z_i\lVert < \lVert\mathbf y-\mathbf z_j\lVert, j=1,...,m, j\neq i,
	\end{eqnarray}
\end{definition}\noindent
where $\{\mathbf z_i\}^m_{i=1} \subset \Re^N$ are referred to as  generators. $\mathbb V_i$ is referred to as a Voronoi tessellation or Voronoi diagram.
\begin{definition} Discrete centroidal Voronoi tessellation. 
	A centroidal Voronoi tessellation (CVT) is a Voronoi tessellation where $\{\mathbf z_i\}$ is also the mass centroid of its Voronoi region. By choosing a uniform density function, we can obtain $\{\mathbf z_i\}$ for the discrete form of CVT \cite{valette2004approximated}\cite{du1999centroidal} as 
	\begin{eqnarray}
		\mathbf z_i = \frac{\sum_{\tau \in \mathbb V_i} \zeta_c(\tau)\zeta_a(\tau)}{\sum_{\tau \in \mathbb V_i} \zeta_a(\tau)},
		\label{eq:cog}
	\end{eqnarray}
	where $\zeta_c(\tau)$ and $\zeta_a(\tau)$ are the functions to obtain the centroid and the area of the face $\tau$, respectively. Centroidal Voronoi diagrams minimize the following energy function
	\begin{eqnarray}
		E_{l2} = \sum^{m}_{i=1}\sum_{\tau \in \mathbb V_i}(\zeta_a(\tau)\lVert\zeta_c(\tau)-\mathbf z_i\lVert^2)
		\label{eq:discrete_energy}
	\end{eqnarray}
\end{definition}\noindent
\begin{definition}Discrete constrained centroidal Voronoi diagram.
	$\{\mathbb V_i\}^m_{i=1}$ is the CCVT of $\mathbb W$ if and only if $\{\mathbf z_i\}^m_{i=1}$ is the constrained mass centroid of those regions. The constrained centroid $\mathbf z_i^c$ of $\mathbb V_i$ on a continuous compact set $\mathbf S \subset \Re^N$ is determined as the projection of $\mathbf z_i$ onto faces in $\mathbf S$ along the proxy normal of $\mathbb V_i$. For a a discrete set of face centroids of $\mathbb M$, this projection is approximately determined as
	\begin{eqnarray}
		min_{\mathbf z_i^c \in \mathbb V_i}(\lVert\mathbf z_i-\mathbf z_i^c\lVert^2)
		\label{eq:cog_prj}
	\end{eqnarray}
\end{definition}
Various energy functions have been proposed for different applications. A $\mathcal{L}^{2,1}$ metric is proposed for capturing anisotropy and segmentation \cite{cohen2004variational}. By combining metrics from spherical and hyperbolic spaces, a unified framework in universal covering space \cite{rong2011centroidal} is proposed to get uniform partitions and high-quality remeshing results. However, these approaches do not align with our scope, leading us to introduce a new energy function that combines distance and normal costs. In the CVT formulation for 3D surfaces, the GD is a natural choice \cite{liu2009centroidal}, but its calculation can be computationally expensive, leading to the use of the Euclidean metric as an approximation. This approximation, however, may result in significant errors on high-curvature surfaces \cite{valette2004approximated}. 
Compared to the $l_2$ norm, the $l_1$ norm is preferable in high dimensional spaces and has been efficiently utilized to approximate the GD in various fields \cite{Mozerov2017Improved}\cite{Briggs2009Audio}. Motivated by this, we select the $l_1$ metric for evaluating the distances on 3D surfaces in the distance cost. The normal cost is used to restrict the variation of triangle normals. The proposed energy function is formulated as follows:
\begin{align}
	\begin{split}
		E = \sum^{m}_{i=1}\sum_{\tau \in \mathbb V_i}\xi(\mathbf z_i, \tau)
	\end{split}
	\label{eq:discrete_energy_total}
\end{align}
where the cost function $\xi$ is defined as
\begin{align}
	\xi(\mathbf z_i, \tau)=(\alpha^{-1}_1\alpha_2\zeta_a(\tau)\lVert\zeta_c(\tau)-\mathbf z_i\lVert_1 + \overline{\alpha}_2\zeta_a(\tau)\Upsilon_n),
	\label{eq:energy_cost}
\end{align}
where $\alpha_1$ is a positive constant used to normalize the first term. It is determined based on the length of the diagonal bounding box of the mesh. $\alpha_2$ is selected within the range of $\in [0,1]$,  $\overline{\alpha}_2 = 1-\alpha_2$. $\Upsilon_n$ is the normal cost and is determined as
\begin{eqnarray}
	\Upsilon_n = \beta(\zeta_n(\tau), \zeta_n(\mathbf z_i))(1-\zeta_n(\tau)\boldsymbol{\cdot} \zeta_n(\mathbf z_i))/2
\end{eqnarray}
where $\zeta_n(\tau)$ and $\zeta_n(\mathbf z_i)$ are the normal vector of face $\tau$ and the proxy normal of $\mathbb V_i$, respectively. The function $\beta$ is selected as $\beta = 1$ if the dot product $\zeta_n(\tau)\boldsymbol{\cdot} \zeta_n(\mathbf z_i)$ $ > \alpha_3$, and $\beta = \alpha_4$ otherwise. $\alpha_3 \in (-1,1)$ and $\alpha_4 > 1 $ are tuning constants. $\alpha_3$ is the threshold at which a higher weight ($\alpha_4$) is applied to the normal cost, aiming to remove and reorganize triangles from the assigned cluster if they exhibit large normal differences compared to the corresponding cluster's proxy normal. 
The CCVT is computed based on the trade-off between distance and normal costs, as described in \eqref{eq:discrete_energy_total}. For meshes with high curvature surfaces, we mitigate the influence of the normal cost to avoid multiple components per cluster, thereby simplifying the post-processing step \cite{valette2004approximated}. This can be achieved by increasing $\alpha_2$ and decreasing $\alpha_3$ and $\alpha_4$. Conversely, for meshes with low curvature surfaces, we decrease $\alpha_2$ and increase $\alpha_3$ and $\alpha_4$ to obtain low curvature clusters.

The CCVT is computed by the Lloyd algorithm as follows. (i) Randomly select $m$ points $\mathbf z_i \in \mathbb W$. (ii) Update Voronoi regions: a face in $\mathbb F$, with the associated centroid $\{\mathbf y_v\}^{n_t}_{v=1}$, is assigned to $\mathbb V_q, q=1,...,m$ if
\begin{align}
	\mathbf z_q = \operatorname*{argmin}_{\mathbf z_i \in \{\mathbf z_i\}^m_{i=1}} \xi(\mathbf z_i, \mathbf y_v)
\end{align}\noindent
(iii) Update the mass centroids of the computed CCVTs using \eqref{eq:cog}. (iv) Update $\mathbf z_i$ to the constrained mass centroid of the CCVT $\mathbb V_i$ using \eqref{eq:cog_prj}. (v) Check if the convergence criteria are met; otherwise, repeat step (ii).

\subsection{Determination of coverage path using the geodesic metric}
The cleaning task is performed by aligning the nozzle axis at $m$ points located at $\mathbf z_i$, normal direction $-\zeta_n(\mathbf z_i$), and the distance from the nozzle to $\mathbf z_i$ is $r_s$. The CPP of this task is to find the shortest path traversing all $\mathbf z_i$. The distances between $\mathbf z_i$ on the mesh $\mathbb M$ are computed using the geodesic metric. However, computing GDs on the entire $\mathbb M$ is computationally intensive and impractical, especially on high resolution meshes. To tackle this issue, we propose a new approach to decompose the computing on the whole mesh into the computing on $m$ subgraphs constructed by edges of each $\mathbb V_i$ and its neighbor-connected clusters as Algorithm \ref{alg_cpp}, Lines 1-7. Here, $\texttt{\textbf{GetAdj}}(\cdot)$ and $\epsilon(\cdot)$ are used to get neighbor-connected clusters ($\cdot$) and edges of the cluster ($\cdot$), respectively. The exact GD and its path are computed using $\texttt{\textbf{ExactGeodesic}}$ \cite{mitchell1987discrete}. $\mathbb C_d$ and $\mathbb P_d$ are dictionary mappings containing cost and tour between generators obtained from computing the shortest path using Dijkstra's algorithm for multiple sources and destinations based on the graph $\mathcal{G}_g$. It is noted that Dijkstra's algorithm is only used to compute distances and paths of non-existing edges in $\mathbb E_g$. Lastly, the coverage path is determined by finding the shortest path to visit each generator $\mathbf z_i$ exactly once. This problem is formulated as the Traveling Salesman Problem (TSP), which is NP-hard. It is probably impossible to find optimal solutions in polynomial time \cite{2opt_approx}. Based on the obtained GD cost in $\mathbb E_g$, the 3-Opt algorithm (Algorithm \ref{alg_cpp}, Line 15) is applied to find the near-optimal solution of the TSP in 3D space. As a result, we obtain a near-optimal tour of $\{\overline{\mathbf z}_i\}^m_{i=1}$ over $\mathbb M$ and the corresponding geodesic path, which represents the coverage path, extracted from $\mathcal P_{geo}$. The notations $\mathbb X \xleftarrow{+} x$ and $\mathbb X \leftarrow x$ represent appending the element $x$ to the list $\mathbb X$ and assigning $x$ to the list $\mathbb X$, respectively. $\mathbb C_d[i,j]$ returns the cost length from $\mathbf z_i$ to $\mathbf z_j$. $\mathbb P_d[i,j]$ is used to get geodesic path between $\mathbf z_i$ and $\mathbf z_j$.
\newline\noindent
\textbf{Complexity.}
Assuming that the processed triangular mesh $\mathbb M$ is closed. The relation between the number of faces and edges $n_e$ \cite{allan1989methodology} is $n_e = 3/2n_t$. The complexity of an exact geodesic algorithm is $\mathcal{O}(n_e^2log(n_e))$ \cite{mitchell1987discrete}. To compute the coverage path, GDs between cluster centroids need to be computed (Algorithm \ref{alg_cpp}, Lines 2-7). Assuming that each cluster has $\varpi$ neighbors on average, our method achieves the complexity  of $\mathcal{O}(m(\varpi/2)((\varpi+1)2n_e/(3m))^2log((\varpi+1)2n_e/(3m))$ 
and requires $\mathcal{O}((\varpi+1)2n_e/(3m))$ space. Without using our proposed method, those distances are determined by computing on the entire edges of $\mathbb M$ with the complexity is $\mathcal{O}((m(m-1)/2)n^2_elog(n_e))$ and requires $\mathcal{O}(n^2_e)$ space, which requires much larger space and is more computationally expensive compared to the proposed decomposition calculation, especially for high resolution meshes. The complexity for the rest computations are $\mathcal{O}(m^3)$(Line 8, Dijkstra) + $\mathcal{O}(m(m-1)/2)$ (Lines 9-14) + $\mathcal{O}(m^3)$(3Opt). Thus, the proposed CPP can be obtained within a reasonable time.
\begin{algorithm}[t]
	\caption{Computation of a coverage path.}
	\label{alg_cpp}
	\SetKwFunction{algo}{algo}\SetKwFunction{proc}{proc}
	\SetKwProg{myalg}{$\texttt{GetCoveragePath}(\{\mathbb V_i\}^m_{i=1}, \{\mathbf z_i\}^m_{i=1})$}{}{}
	\myalg{}{
		\nl $\mathcal{P}_{geo} = \emptyset$;$\mathbb E_g = \emptyset$;\\
		\nl \For{$i=1$ \KwTo $m$}{ 
			\nl $\mathbb V_a, \mathbf z_a = \texttt{\textbf{GetAdj}}(\mathbb V_i$); $\mathbb E_p = \epsilon(\mathbb V_i \cup V_a)$;\\
			\nl \For{$\mathbf z$ \textup{\textbf{in}} $\mathbf z_a$}{
				\nl \If{\textup{(}$\mathbf z_i, \mathbf z_a$\textup{)} \textup{\textbf{not in}} $\mathbb E_g$}{ 
					\nl$c, p = \texttt{\textbf{ExactGeodesic}}(\mathbb E_p,\mathbf z_i, \mathbf z_a)$;\\
					\nl $\mathbb E_g \xleftarrow{+} ((\mathbf z_i, \mathbf z_a),c); \mathcal{P}_{geo} \xleftarrow{+} p $;\\
				}
			}
		}
		\nl $\mathbb N_g = \{\mathbf z_i\}^m_{i=1};\mathcal{G}_g = (\mathbb N_g,\mathbb E_g)$; \\
		\nl $\mathbb C_d, \mathbb P_d = \texttt{\textbf{Dijkstra}}(\mathcal{G}_g)$;\\
		\nl \For{$i=1$ \KwTo $m$}{ 
			\nl \For{$j=i+1$ \KwTo $m$}{ 
				\nl \If{\textup{(}$\mathbf z_i, \mathbf z_j$\textup{)} \textup{\textbf{not in}} $\mathbb E_g$}{ 
					\nl $\mathbb E_g \xleftarrow{+} ((\mathbf z_i, \mathbf z_j),\mathbb C_d[i,j])$;\\ 
					\nl $\mathcal{P}_{geo} \xleftarrow{+} P_d[i,j] $;\\
				}
			}
		}
		\nl \KwRet $\textbf{3Opt\_GeoPath}(\mathbb E_g, \mathcal{P}_{geo})$;}
\end{algorithm} 
\subsection{Correction of Infeasible Robot Configurations and Computation of Final Robot Trajectory}\hfill\\
\begin{figure}[t]\centering
	\includegraphics[width=0.47\textwidth]{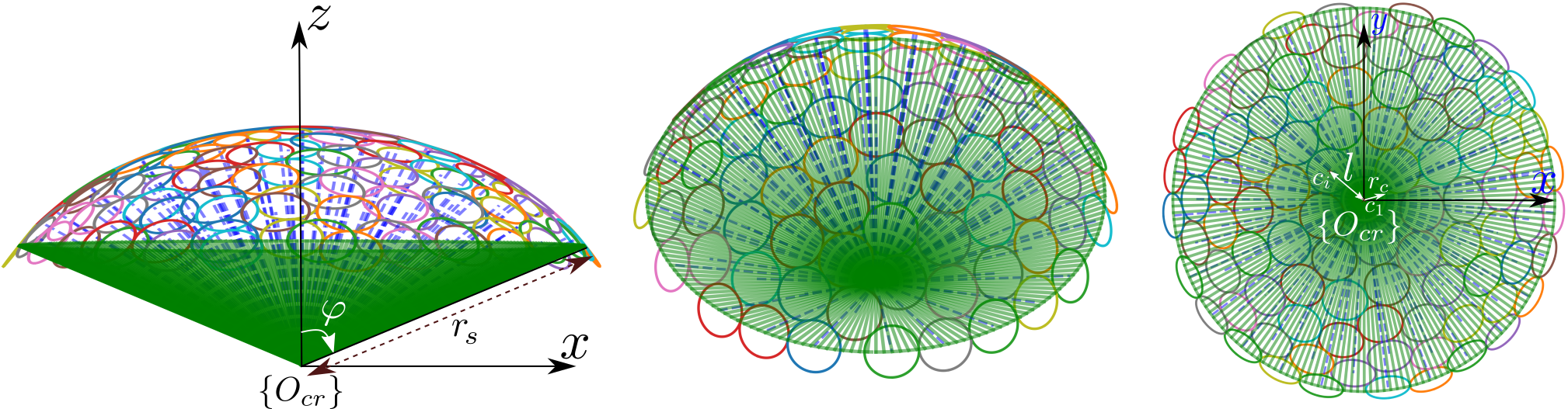}
	\caption{\label{fig_cr}Set of candidate rays in different views.}
\end{figure}
Let $\mathbf{r}_{i} =\{\overline{\mathbf z}_i,\zeta_n(\overline{\mathbf z}_i)\}$, $\zeta_n(\overline{\mathbf z}_i)=[n_{cx}, n_{cy}, n_{cz}]^T$, and $\mathbb R_{opt} = \{\mathbf{r}_{i}\}^m_1$. 
For the cleaning task, it is crucial that the ray $\mathbf{r}_i$ has no obstructions, enabling unobstructed airflow from the nozzle to $\overline{\mathbf{z}}_i$ (or unobstructed camera view for model-based inspection or 3D reconstruction tasks in regions of interest). However, this property cannot be guaranteed solely based on the obtained proxy normals $\zeta_n(\overline{\mathbf{z}}_i)$.
 Also, there exists a robot configuration,  i.e., an inverse kinematics (IK) solution, ensuring the tool is aligned to $\mathbf r_i$ without colliding with the surrounding environment. We propose an optimization problem (see appendix A) to construct a set of candidate rays representing the possible directions of the nozzle at $\overline{\mathbf z}_i$ as shown in Fig. \ref{fig_cr}. The obtained result is a set $\mathbb{C}_{cr}= \{c_1, c_2,...,c_{N_c}\}$ containing $N_c$ center points of circles of radius $r_c$ (nozzle radius). 
These circles are (almost) uniformly distributed in a specified region, defined by an elevation angle $\varphi$, of a sphere surface of radius $r_s$. Also, the returned centers of $\mathbb{C}_{cr}$ are then sorted in ascending order of distance to $c_1$. 
The set of near-optimal, collision-free IKs is computed using Algorithm \ref{alg_cr}. Those IKs are used to compute coverage trajectories to perform the cleaning task. Since the rotation about the ray is irrelevant for cleaning tasks, we can discretize the rotation about each ray into a set of angles $\Theta$ as
\begin{equation}
	\Theta = \{2\pi(i-1)/I\}, i = 1,..,I, 
\end{equation}
where $I$ is the discretization step size, thus, for each $\mathbf r_i$, there are $I$ possible configurations. We use OpenRAVE \cite{openrave} for computing IKs, checking collisions between the corresponding robot configurations and environment ($\texttt{\textbf{getIKs}}$ and $\textup{\texttt{\textbf{isValidCfg}}}$, Algorithm \ref{alg_cr}), and performing motion planning to obtain final robot trajectory. A valid robot configuration, i.e., $\textup{\texttt{\textbf{isValidCfg}}}(\mathbf{r}, \theta_r)$ returns \texttt{True} if ray $\mathbf{r}$ has an elevation angle less than $\theta_r$ and the obtained configuration is collision-free with the surroundings.	
Trimesh \cite{trimesh} library is used to check self-collision $\mathbf{r}_{i} \cap \mathbb{F}$, Algorithm \ref{alg_cr}. A ray $\textbf{r}_x$ is considered invalid if it either collides with the mesh or has no IK solution aligned with that ray. 
An invalid ray is corrected (if possible) by using $\texttt{\textbf{getIKs}}$ and aligning the candidate rays in $\texttt{\textbf{AlignCR}}$ to new poses such that $O_{cr}$ is located at $\overline{\mathbf z}_i$ and $c_1$ lies in the direction of $\textbf{r}_x$.
A set of valid robot configurations $\mathcal{C}_{free}$ is obtained by applying Algorithm \ref{alg_cr}, Lines 1-7. $\overrightarrow{\overline{\mathbf z}_xo_k}$ is the normalized vector from $\overline{\mathbf z}_x$ to $o_k$. $\mathcal R_a(b) \in SO(3)$ is the rotation matrix representing the rotation about the $a$-axis by angle $b$. $\mathcal T(b) \in SE(3)$ is the translation matrix translating a point from the origin to position $b$.
Based on the collision-free set of IKs $\mathcal{C}_{free}$, an undirected graph $\mathbb{G}$ is constructed to find a set of optimal robot configurations in configuration space, where the tour cost has a minimal length in the configuration-space metric. An optimal tour of IK solutions over coverage lines $\overline{\mathcal{C}}_{free}$ can be obtained using the function $\textup{\texttt{\textbf{ComputeOptCfg}}}(\mathcal{C}_{free})$, Algorithm \ref{alg_cr} (by solving $\mathbb{G}$ using Dijkstra's algorithm in the same manner as \cite{suarez2018robotsp}).
\begin{algorithm}[t]
	\caption{Determination of near-optimal, valid, and collision-free robot configurations.}
	\label{alg_cr}
	\SetKwFunction{algo}{algo}\SetKwFunction{proc}{proc}
	\SetKwProg{myalg}{$\texttt{GetOptRobotCfgs}(\mathbb R_{opt}, \mathbb{C}_{cr})$}{}{}
	\myalg{}{
		\nl$\mathcal{C}_{free} = \emptyset$;\\
		\nl \For{$i=1$ \KwTo $m$}{ 
			\nl \uIf{$\mathbf{r}_{i} \cap \mathbb{F} == \emptyset$ \textup{\textbf {and}} $\textup{\texttt{\textbf{isValidCfg}}}(\mathbf r_i, \theta_r)$}{
				\nl$\mathcal{C}_{free} \xleftarrow{+} \textup{\texttt{\textbf{getIKs}}}(\mathbf r_i)$;\\
			}
			\uElse{
				$\mathbf r_t = \textup{\texttt{\textbf{GetFreeRay}}}(\mathbf{r}_{i}, \mathbb{C}_{cr})$; \\
				\nl \If{$\mathbf r_t $ $\neq$ $ \mathbf{None}$}{
					\nl$\mathcal{C}_{free} \xleftarrow{+} \textup{\texttt{\textbf{getIKs}}}(\mathbf r_t)$;\\
				}
			}
		}
		\nl \KwRet $\textup{\texttt{\textbf{ComputeOptCfg}}}(\mathcal{C}_{free})$;
	}
	\SetKwProg{myalg}{$\texttt{GetFreeRay}(\mathbf{r}_{x}, \mathbb{C}_{cr})$}{}{}
	\myalg{}{
		\nl $\overline{\mathbb{C}}_{cr} = \texttt{\textbf{AlignCR}}(\mathbf{r}_{x}, \mathbb{C}_{cr})$\\
		\nl \For{$k=1$ \KwTo $N_c$}{ 
			\nl $o_k = \overline{\mathbb{C}}_{cr}^{[k]}; \mathbf{r}_{c} = \{\overline{\mathbf z}_x,\overrightarrow{\overline{\mathbf z}_xo_k}\}$;\\
			\nl \If{$\mathbf{r}_{c} \cap \mathbb{F} == \emptyset$ \textup{\textbf {and}} $\textup{\texttt{\textbf{isValidCfg}}}(\mathbf{r}_{c},\theta_r)$}{
				\nl \KwRet $\mathbf{r}_{c}$;
			}
		}
		\nl \KwRet $\mathbf{None}$;}
	\SetKwProg{myproc}{$\texttt{\textbf{AlignCR}}(\mathbf{r}_{x}, \mathbb{C}_{cr})$}{}{}
	\myproc{}{
		\nl     $\phi_y = \text{arccos}(n_{cz}); \phi_z = \text{arctan2}(n_{cy},n_{cx})$;\\
		\nl    $\mathbb C_{tmp} \leftarrow \emptyset$; $\mathcal{T}_a = \mathcal{T}(\overline{\mathbf z}_x)\mathcal{R}_z(\phi_z)\mathcal{R}_y(\phi_y)$;\\
		\nl \For{$k=1$ \KwTo $N_c$}{
			\nl $\mathbb C_{tmp} \xleftarrow{+} \mathcal{T}_a \mathbb{C}_{cr}^{[k]}$;\\
		}
		\nl \KwRet $\mathbb C_{tmp}$;}
\end{algorithm} 
\section{Experiments}
\begin{table*}
	\centering
	\caption{Comparison results.}
	\label{table1}
	\begin{tabular}{|c|c|c|c|c|c|c|c|c|c|c|c|c|c|c|c|c|} 
		\hline
		& \multicolumn{4}{c|}{Coverage (\%)} & \multicolumn{4}{c|}{Overlap (\%)} &  \multicolumn{4}{c|}{RSD ($\%$)}&   \multicolumn{4}{c|}{$\mathbb F_{\textup{unreach}}$ (\%)}\\ 
		\hline
		& $bl$ & $bl_{\times3}$ & $l_2$ & Ours                    & $bl$ & $bl_{\times3}$ & $l_2$ & Ours               & $bl$ & $bl_{\times3}$ & $l_2$ & Ours                                & $bl$ & $bl_{\times3}$ & $l_2$ & Ours               \\ 
		\hline
		Bunny & 64.6 & 95.1 & 97.7  & 97.6                & 44.1 & 86.3  & 10.7  & 11.5                   & 0.8 & 0.43  & 1.1  & 1.3                                      & 5.7 & 1.6  & 5.4   & 0.5                     \\ 
		\hline
		Deer  & 80.6 & 97.6 & 98.9  & 97.5               & 63.6 & 92.7  & 26.9  & 28.8                  & 2.1 & 1.2  & 2.2  & 2.4                                      & 21.3 & 5.9  & 21.3  & 5.9                    \\ 
		\hline
		Nene & 67.3 & 95.3 & 94.4  & 92.3                & 45.1 & 87.9  & 10.7  & 8.3                   & 1.4 & 0.86  & 2.0  & 2.4                                      & 16.7 & 5.8  & 15.0  & 2.5                    \\ 
		\hline
		Lion  & 69.0 & 93.6 & 95.1  & 92.4               & 57.8 & 86.1  & 11.3  & 11.4                  & 2.1 & 1.2  & 1.6  & 1.9                                      & 19.7 & 5.1  & 17.7  & 13.7                 \\
		\hline
	\end{tabular}
\end{table*}
We use a Ubuntu workstation with an Intel Xeon$^{\circledR}$ W-2255 CPU, 3.70GHz x 20, 24-GB RAM to benchmark the proposed algorithm on four objects with the order ascending of complexity and size (from left to right in Fig. \ref{fig_obj_res}), including Bunny (69,451 faces), Deer (139,306 faces), Nene (193,254 faces), and Lion (206,328 faces). 
The parameters are selected as $\alpha_1 = 1/6d_{dgn}, \alpha_2 = 0.93$, $\alpha_4 = 7$, where $d_{dgn}$ is the diagonal length of the object bounding box. We select $\alpha_3 = 1/1.9$ for Bunny, Deer, and Nene. For Lion, we select $\alpha_3 = 1/3$ due to the high roughness of its geometry. The number of clusters $m$ is determined based on the nozzle radius $r_c = 5\sqrt{2}\times10^{-3}$ m, i.e., $m = \zeta_a(\mathbb F)/\sigma_e$, where $\zeta_a(\mathbb F)$ is the total area of all triangle faces and $\sigma_e = \pi r_c^2$ is the expected cleaning area at a robot configuration.
\subsection{Coverage Path Planning Results}
\begin{figure}[t]\centering
	\includegraphics[width=0.36\textwidth]{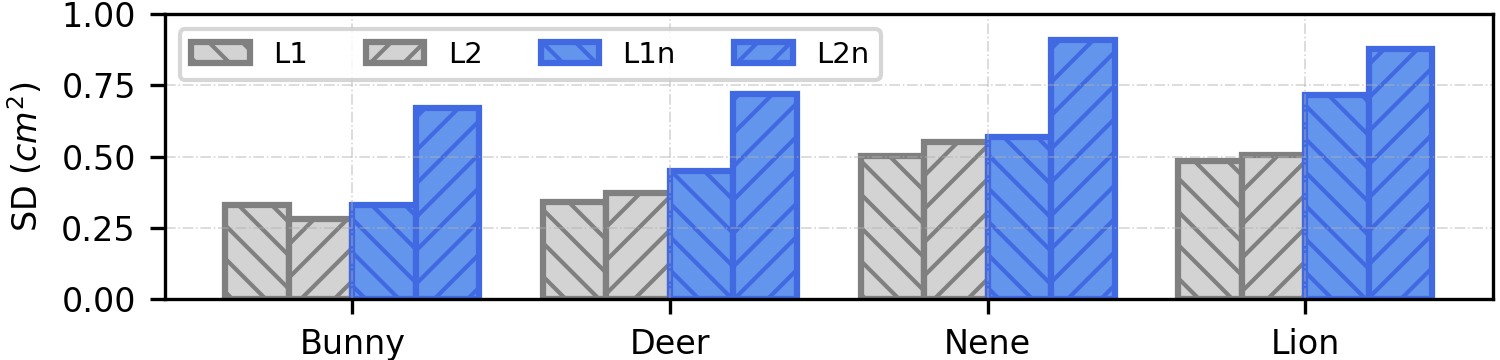}
	\caption{\label{fig_area_std} The standard deviations of the cluster areas of the four objects.}
\end{figure}

\begin{figure}[t]\centering
	\includegraphics[width=0.37\textwidth]{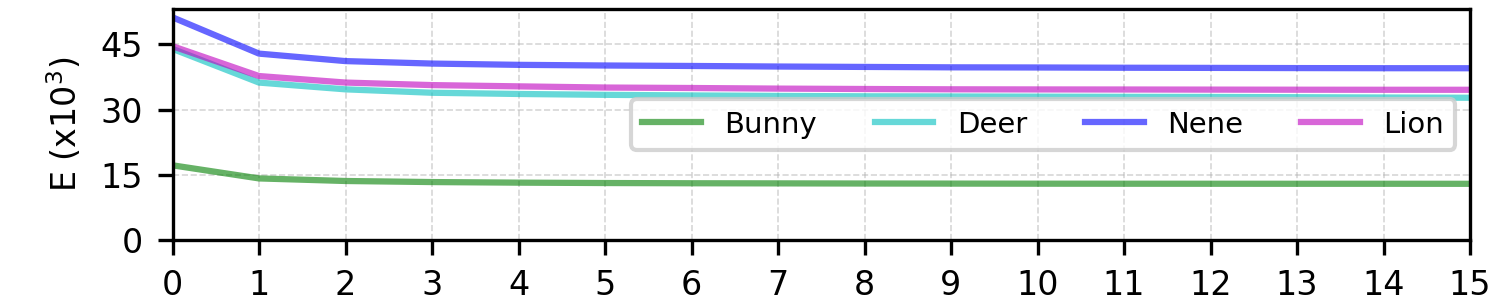}
	\caption{\label{fig_energy} Convergence of the energy function of the four objects.}
\end{figure}

\begin{figure}[t]\centering
	\includegraphics[width=0.43\textwidth]{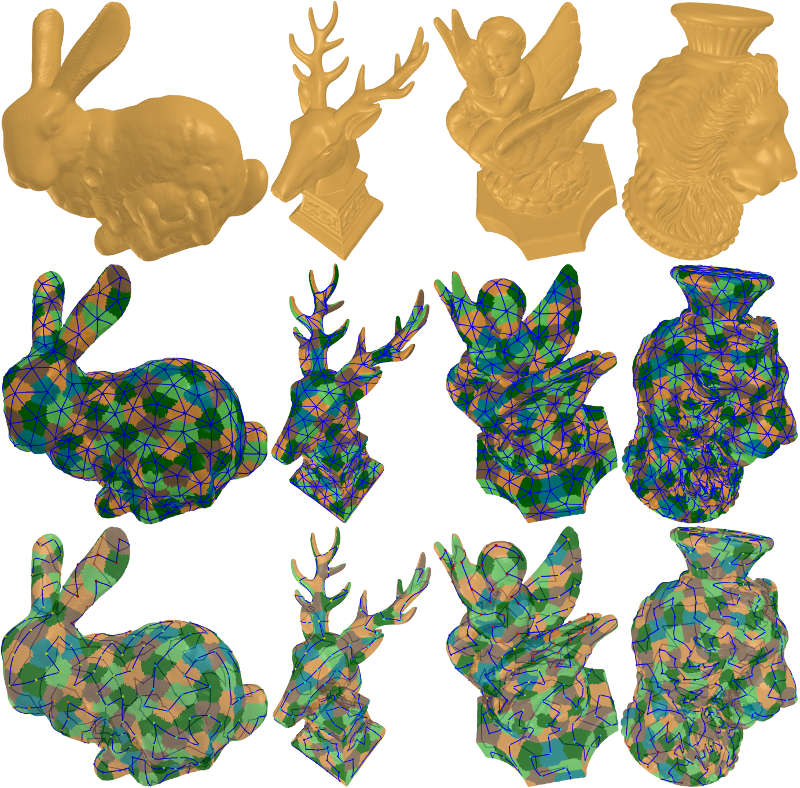}
	\caption{\label{fig_obj_res} Results of the planned coverage paths (Algorithm \ref{alg_cpp}).}
\end{figure}

To assess the effectiveness of the proposed energy function in generating uniform cluster areas, we perform mesh segmentation experiments using different norm types in the distance function, represented by \eqref{eq:discrete_energy} with $l_1$-norm ($l_1$), \eqref{eq:discrete_energy} with $l_2$-norm ($l_2$), \eqref{eq:discrete_energy_total} with $l_1$-norm ($l_{1n}$, proposed method), and \eqref{eq:discrete_energy_total} with $l_2$-norm ($l_{2n}$). We evaluate its uniformity by computing the SD of the obtained cluster areas for the four objects, as shown in Fig. \ref{fig_area_std}. Results show that $l_2$ yields the lowest SD in the Bunny object (lowest curvature). For higher curvature meshes, however, $l_1$-based norms perform better at generating more uniform cluster areas. Particularly, the use of $l_{1n}$ with the normal cost results in significantly more uniform cluster areas compared to $l_{2n}$. Fig. \ref{fig_energy} shows the convergence of the energy function $l_{1n}$ of all objects after 16 iterations. The result of computing CCVTs is described as the second row of Fig. \ref{fig_obj_res}. The blue lines are geodesic paths connecting all generators of neighbor-connected clusters. The triangle normals on each cluster vary within a specific range while maintaining a nearly uniform distribution of clusters over the mesh. 

The CPs on the meshes are shown in the third row of Fig. \ref{fig_obj_res}, where a unique, shortest path connects all generators to form CPs on mesh models. The computation time of our CPP mainly depends on the computation of exact GDs (Algorithm \ref{alg_cpp}, Lines 2-7). Using the proposed decomposition method, the GDs between segmented clusters for the four objects are computed in 10 s (271 clusters), 23 s (209 clusters), 37 s (233 clusters), and 33 s (332 clusters), respectively. This is a significant improvement compared to the computing times of 39,240 s, 56,160 s, 91,908 s, and 225,828 s without applying the decomposition approach. The proposed method results in a slightly longer computation time for Nene than Lion due to Lion's more complex geometry, which requires a larger number of segmented clusters.

We further compare our proposed method ($l_{1n}$) with a conventional CCVT approach ($l_2$) and a baseline \cite{jing2017model} by additional CPP criteria as Table \ref{table1}. $bl$ and $bl_{\times3}$ are the results produced by the baseline method using the same number of VPs and three times the number of VPs generated by our method, respectively.
Here, we use GD from the triangles of interest to the generators of adjacent clusters and a threshold determined by $r_c$ to estimate coverage and overlap rates. RSD is the relative standard deviation of the expected area per cluster computed based on clusters' area and $\sigma_e$. $\mathbb F_{\textup{unreach}}$ is the set of unreachable triangles per object in terms of the normal cost. A triangle is added to $\mathbb F_{\textup{unreach}}$ if the angle constructed by its normal and the corresponding generator's normal is larger than a certain value $\theta_0$, i.e., $\theta = \arccos({\zeta_n(\tau)\boldsymbol{\cdot}\zeta_n(\mathbf z_i)}) > \theta_0$. The physical meaning of $\mathbb F_{\textup{unreach}}$ is that if $\theta$ is greater than that value, then the performance of the task will be degraded. For the cleaning task, we select $\theta_0 = \pi/3$ rad. 
For each object, $\mathbb F_{\textup{unreach}}$ (\%) is determined by $\mathbb F_{\textup{unreach}}/n_t$. 
Table \ref{table1} shows that the baseline method $bl$ has significantly higher rates of overlapping and unreachable features ($\mathbb F_{\textup{unreach}}$) and significantly lower coverage rates compared to our proposed method ($l_{1n}$). The coverage rates can be improved by $bl_{\times3}$; however, the overlapping becomes markedly worse. Both $l_2$ and our proposed method ($l_{1n}$) achieve nearly complete coverage with much lower overlapping rates compared to the baseline.

\begin{figure}[t]\centering
	\includegraphics[width=0.49\textwidth]{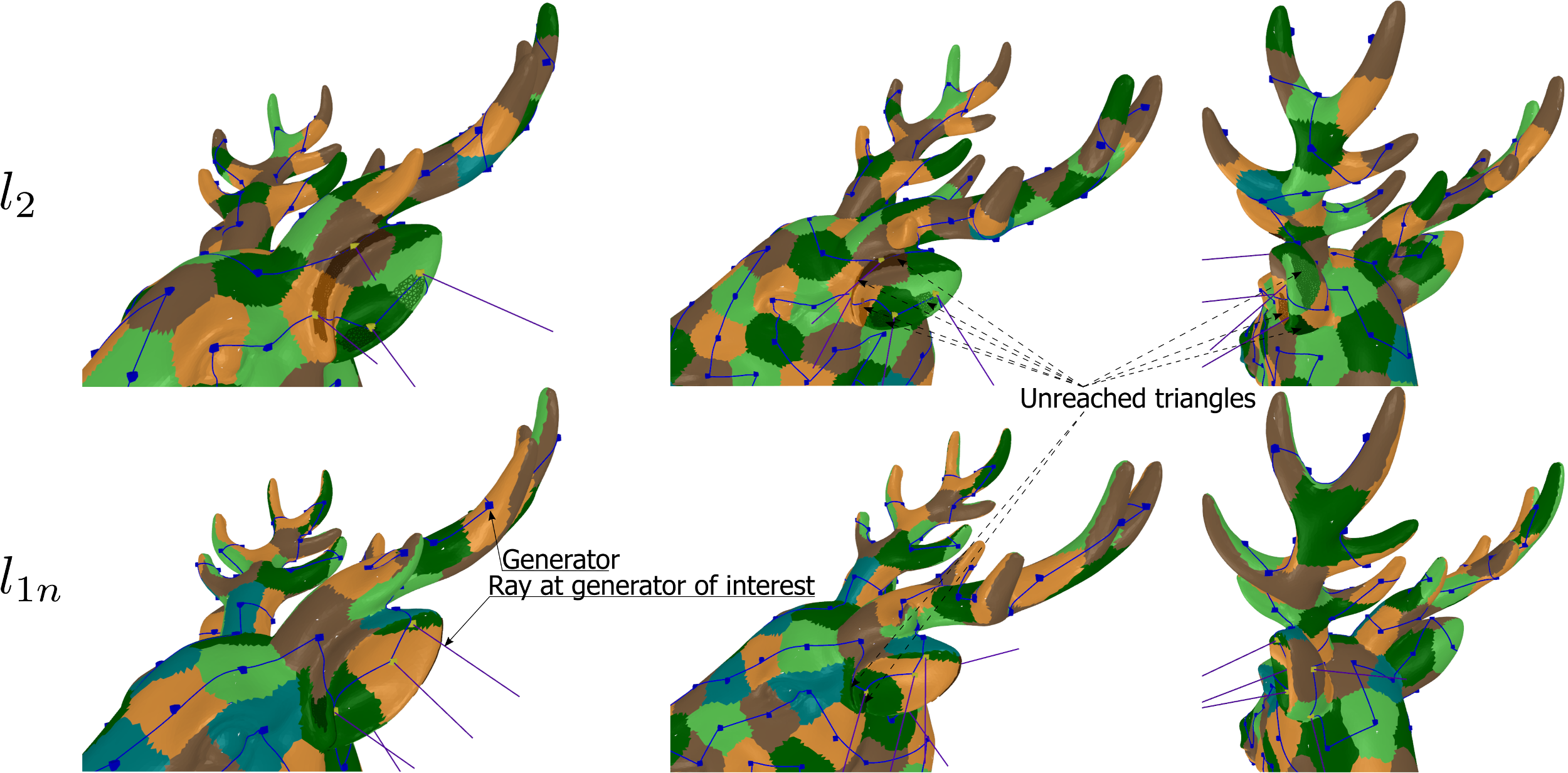}
	\caption{\label{fig_antlers} CCVT results in different views around the left ear of the Deer.}
\end{figure}

Fig. \ref{fig_antlers} shows the computed CCVTs in different views around the Deer's left ear. Generators on the ear are marked as yellow. For each cluster, edges of unreached triangles are visualized in darker colors compared to its corresponding cluster color. By using the proposed energy function, our method achieves a significantly lower portion of unreached elements. The presence of larger and more numerous dark portions (high $\mathbb F_{\textup{unreach}}$ rate) in the $l_2$ method degrades the performance of the cleaning task as the airflow's impact on the target clusters is either diminished or absent. For model-based inspection and 3D reconstruction tasks, capturing entire clusters at computed VPs is not feasible, rendering VP planning using a set covering problem \cite{jing2017model} ineffective. With nearly complete coverage, considerably low rates of RSD and $\mathbb F_{\textup{unreach}}$, our proposed method allows for generating a high-quality set of VPs for surface-based coverage tasks. 

\begin{figure}[t]\centering
	\includegraphics[width=0.49\textwidth]{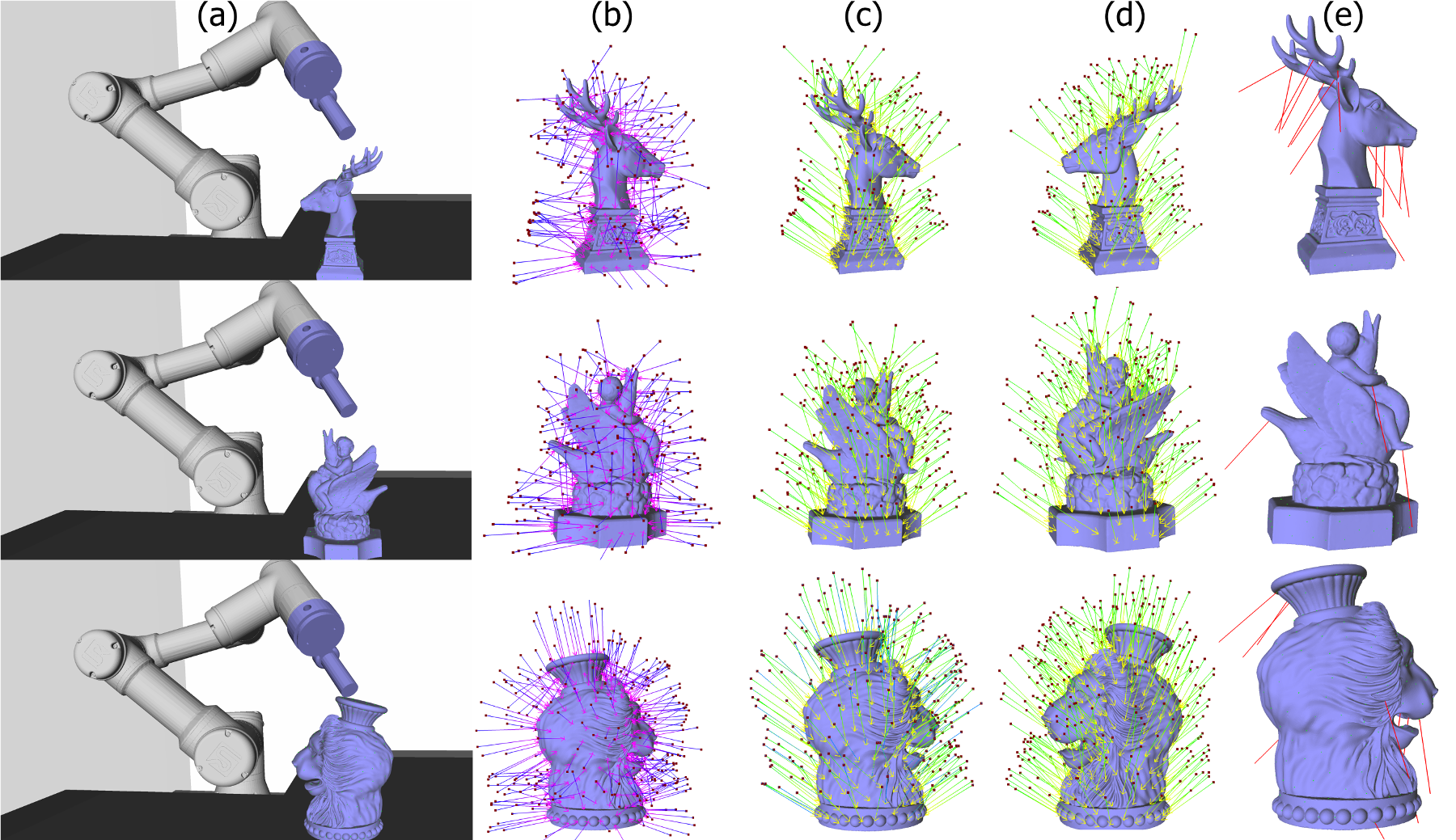}
	\caption{\label{fig_cfg_res} Results of finding the set of valid configurations (Algorithm 2). \newline(a) Object and robot in OpenRAVE. (b) Pre-calculated rays. (c) \& (d) Rays of valid configurations after applying Algorithm 2. (e) Unrecoverable rays.}
\end{figure}
\subsection{Actual Cleaning Demonstration}
\begin{figure*}[t]\centering
	\includegraphics[width=1.0\textwidth]{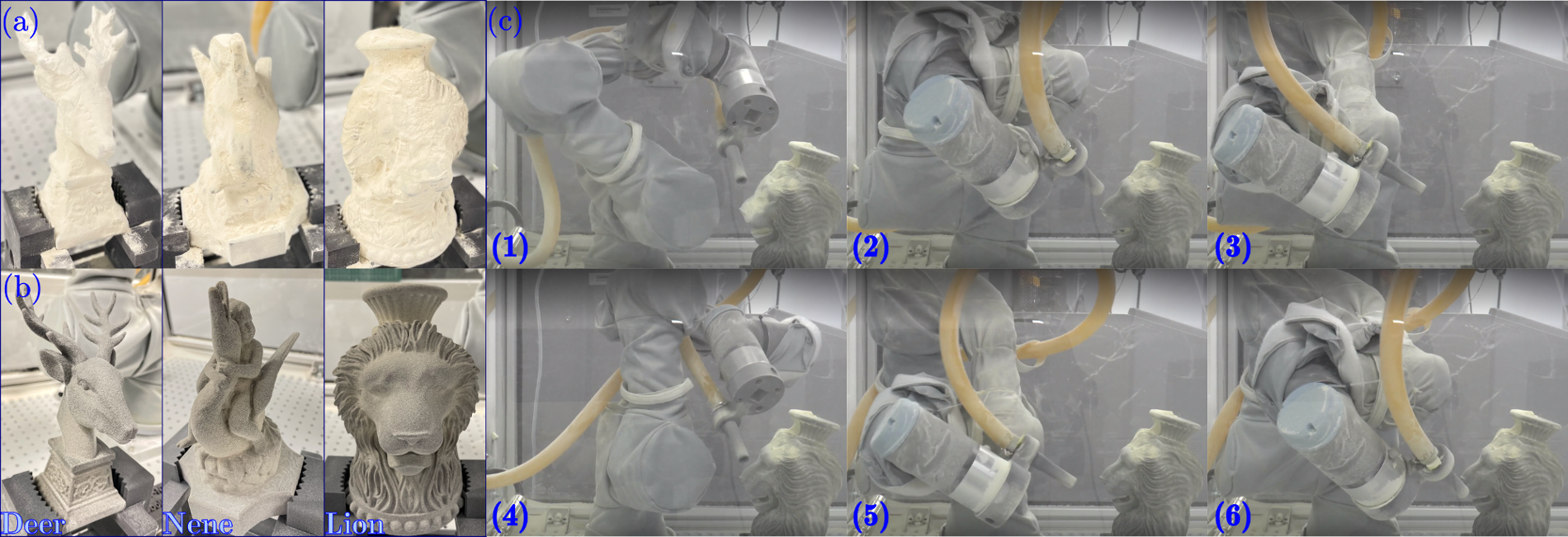}
	\caption{\label{fig_snap} Actual photos of 3DP samples. (a) After unpacking from a 3D printing station. (b) After performing the cleaning process. (c) Snapshots of the actual cleaning actions in an attempt to remove residual powder from Lion's mouth. The full video is available at $\texttt{https://youtu.be/eD5AjJKkEaI}$. }
\end{figure*}
Practical cleaning demonstrations are conducted using a collaborative robot (UR3e) for Deer, Nene, and Lion objects, which have high geometry complexity. The 3DP objects after unpacking from a printing station (HP MJF5200) are shown in Fig. \ref{fig_snap}(a). The blasting pressure is set at 60 psi, $\theta_r$ = $\pi/3$ rad, and $r_s$ = 0.05 m. Fig. \ref{fig_cfg_res} shows the results of finding a set of valid robot configurations using Algorithm 2. The relative position of the object and the UR3e is (0.24 m, 0.2 m, 0.08 m), as shown in Fig. \ref{fig_cfg_res} (a). By using Algorithm 2, a part of the rays in $\mathbb R_{opt}$ (Fig. \ref{fig_cfg_res} (b)) is either adjusted by correcting their normal directions to attain feasible IK solutions $\mathcal{C}_{free}$, as shown in Fig. \ref{fig_cfg_res} (c) and (d), or removed because there is no feasible IK solution satisfying the condition in Algorithm 2, line 5, as shown in Fig. \ref{fig_cfg_res} (e). The set of unrecoverable rays can be reduced by increasing $\theta_r$ depending on the manipulator's arm length and the object placement. However, a larger $\theta_r$ also increases the planning and execution time, as it requires more time for correcting rays and transitioning between highly stretched configurations.
Based on $\overline{\mathcal{C}}_{free}$, the final trajectory is computed using built-in functions in OpenRAVE and is sent to the robot using the Universal Robots ROS Driver. Fig. \ref{fig_snap}(c) shows the sequential snapshots of the cleaning actions for the Lion, where the robot fits its configurations properly to the computed collision-free configuration at generators. As a result, all residual powder is removed from Lion's mouth and also from the entire object surface, which demonstrates the effectiveness of our approach. The cleaning time is about 300 s. Although the computed coverage is not complete, practical implementation achieves nearly complete coverage as the actual affected area of the nozzle surpasses $r_c$.

In depowdering tasks, the residual powder on the object surface undergoes variations as it is blown away during the cleaning process. In addition, the affected cleaning area in specific configurations depends on various factors, such as surface properties (material, geometry complexity), the adhesion profile of unfused powder, and airflow power, making analytical determination of the affected cleaning size per VP impossible. In our approach, we select $r_s$, $\sigma_e$, $\theta_r$, and blasting pressure from practical perspectives to ensure sufficient airflow for effective removal of residual powder at computed robot configurations, while consistently maintaining the distance of $r_s$ at these configurations. However, this distance is relaxed during transitions between configurations to provide flexibility and rapidity for relocating the nozzle and the attached hose.

\section{Conclusion}
In this study, we have introduced a new approach to obtaining a nearly complete coverage path on general surfaces and high-quality VPs using the information extracted from their mesh models. By introducing a new energy function, the mesh model is segmented into multiple clusters using the CCVT method such that coverage quality (uniform cluster areas) and the variation of triangle normals can be obtained harmoniously. Our results showed that using the $l_1$-norm in the proposed energy function outperforms the $l_2$-norm in achieving lower standard deviations of cluster areas on complex surfaces.
A decomposition calculation is proposed to speed up the finding of the shortest path visiting all segmented clusters using the geodesic metric. As a result, we can attain high coverage, low overlapping, and low curvature of segmented clusters on the coverage path within a reasonable time. An effective optimization-based strategy is then proposed to address the self-occlusion of viewpoints and support the finding of valid robot configurations. Then, a collision-free robot trajectory is computed, allowing the CPP to be applied to cleaning and model-based inspection applications. We validated the effectiveness of our approach on an industrial cleaning platform used for additive manufacturing. Our future plans include: (1) Investigating advanced approaches \cite{liu2009centroidal}\cite{rong2010gpu} to accelerate the construction of CCVTs. (2) Implementing faster approaches with high-quality approximated GDs \cite{adikusuma2020fast} to accelerate CPP's computation. (3) Developing advanced model-based inspection and 3D reconstruction algorithms based on the obtained high-quality VPs.

	\appendices
\section*{Appendix}
\textbf{Appendix A. Construct the set of candidate rays using optimization.}

We propose the following optimization problem to construct the set of candidate rays:
\begin{eqnarray}
	\underset{c_i}{\text{minimize}} \quad & \sum_{i=2}^{N_c} x_i^2 +y_i^2\label{eq_cr_opt}\\
	\text{subject to} \quad & (x_1,y_1) = (0,0),\\
	\quad & z_i \geq R\cos(\varphi),\label{eq_cr_cnst1}\\
	\quad & \Vert c_i - c_j\Vert_2 \geq l,\label{eq_cr_cnst2}\\
	\quad & i= 1,...,N_c, j= i+1,...,N_c,\nonumber
\end{eqnarray}
where $\mathbf z_i = \sqrt{R^2 - x_i^2 - y_i^2}$, $c_i=[x_i, y_i, z_i]^T$, $l$ is the distance between two adjacent circles  determined as
\begin{equation}
	l = 2r_c\cos(\arctan(\frac{r_c}{r_s})),
\end{equation}
The objective function (\ref{eq_cr_opt}) and the function on the left-hand side of (\ref{eq_cr_cnst2}) are both convex. Thus, the above formulation is not a convex optimization problem. This problem can be solved using disciplined convex-concave programming \cite{shen2016disciplined}.

\section*{ACKNOWLEDGMENT}
We would like to thank Dr. Yang Liu for sharing insightful comments on the CCVT topic, as well as Daren Ho for his generous support in operating the cleaning platform.

\bibliographystyle{IEEEtran}
\bibliography{IEEEabrv,references}
\end{document}